\documentclass[journal]{IEEEtran}
\usepackage{cite}
\usepackage{xcolor}

\ifCLASSINFOpdf
   \usepackage[pdftex]{graphicx}
   \usepackage{subcaption}
  \graphicspath{{./fig/}}
\else
\fi
\usepackage{amsmath}
\usepackage{array}
\usepackage{multirow}
\usepackage{url}

\begin{document}
%
\title{Feature Pyramid and Hierarchical Boosting Network for Pavement Crack Detection}
%
%
%

\author{Fan~Yang*,
        Lei~Zhang*,
        Sijia Yu, Danil Prokhorov, Xue~Mei, and~Haibin~Ling
\thanks{
	F. Yang, S. Yu, and H. Ling are with the Department
of Computer and Information Sciences, Temple University, Philadelphia,
PA, 19122 USA. e-mail: (fyang@temple.edu; sijia.yu@temple.edu; hbling@temple.edu).}
\thanks{L. Zhang is with Department of Radiology, University of Pittsburgh, Pittsburgh, PA 15213, USA. (e-mail: cszhanglei@gmail.com)}
\thanks{ D. Prokhorov is with the Toyota Research Institute, North America, Ann Arbor, Michigan 48105 USA. (e-mail: danil.prokhorov@toyota.com)}
\thanks{X. Mei is with TuSimple, San Diego, California 92122 USA. (e-mail: nathanmei@gmail.com)}
\thanks{* equal contribution}}

%
%

\markboth{IEEE TRANSACTIONS ON INTELLIGENT TRANSPORTATION SYSTEMS, UNDER REVIEW.}%
{Shell \MakeLowercase{\textit{et al.}}: Bare Demo of IEEEtran.cls for IEEE Journals}
%



\maketitle

\begin{abstract}
Pavement crack detection is a critical task for insuring road safety. 
 Manual crack detection is extremely time-consuming. Therefore, an automatic road crack detection method is required to boost this progress. However, it remains a challenging task due to the intensity inhomogeneity of cracks and complexity of the background, e.g., the low contrast with surrounding pavements and possible shadows with similar intensity. Inspired by recent advances of deep learning in computer vision, 
 we propose a novel network architecture, named Feature Pyramid and Hierarchical Boosting Network (FPHBN), for pavement crack detection.
The proposed network integrates context information to low-level features for crack detection in a feature pyramid way. And, it balances the contributions of both easy and hard samples to loss by nested sample reweighting in a hierarchical way during training.
In addition, we propose a novel measurement for crack detection named average intersection over union (AIU).
To demonstrate the superiority and generalizability of the proposed method, we evaluate it on five crack datasets and compare it with state-of-the-art crack detection, edge detection, and semantic segmentation methods. Extensive experiments show that the proposed method outperforms these methods in terms of accuracy and generalizability. \textcolor{red}{Code and data can be found in \url{https://github.com/fyangneil/pavement-crack-detection}}
\end{abstract}


%
\IEEEpeerreviewmaketitle

\section{Introduction}
%
%
%
%
\IEEEPARstart{C}{rack} is a common pavement distress, which is a potential threat to road and highway safety. To maintain the road in good condition, localizing and fixing the cracks is a vital responsibility for transportation maintenance department. One major step of the task is crack detection. However, manual crack detection is considerably tedious and requires domain expertises. To alleviate the workload of expertisers and facilitate the progress of road inspection, it is necessary to  achieve automatic crack detection.

With the development of technologies in computer vision, numerous efforts have been devoted to applying computer vision technologies to perform automatic crack detection  \cite{zhang2016road,shi2016automatic,huang2006automatic,zou2012cracktree,schmugge2017crack,kaseko1993neural,jahanshahi2013innovative}. In early days, Liu et al \cite{liu2008novel} and Kaseko et al \cite{kaseko1993neural} use threshold-based approaches to find crack regions based on the assumption that real crack pixel is consistently darker than its surroundings.
Currently, most of the methods are based on hand crafted feature and patch-based classification. Many types of such features have been used for crack detection such as: Gabor filters \cite{chanda2014automatic}\cite{medina2014enhanced}, wavelet features \cite{zhou2006wavelet}, Histogram of Oriented Gradient (HOG) \cite{kapela2015asphalt}, and Local Binary Pattern (LBP) \cite{hu2010novel}. These methods encode the local pattern but lack global view  of crack. To conduct crack detection from a global view, some works \cite{amhaz2016automatic,zou2012cracktree,fernandes2014pavement} carry out crack detection by taking into account photometric and geometric characteristics of pavement crack images. These methods partially eliminate noises and enhance continuity of detected cracks.

While these methods perform crack detection in a global view, their detection performance are not superior when dealing with cracks with intensity inhomogeneity or complex topology. The failures of these methods can be attributed to lacking robust feature representation and ignoring interdependency among cracks.
\begin{figure}[!t]
	\centering
	\includegraphics[width=0.5\textwidth]{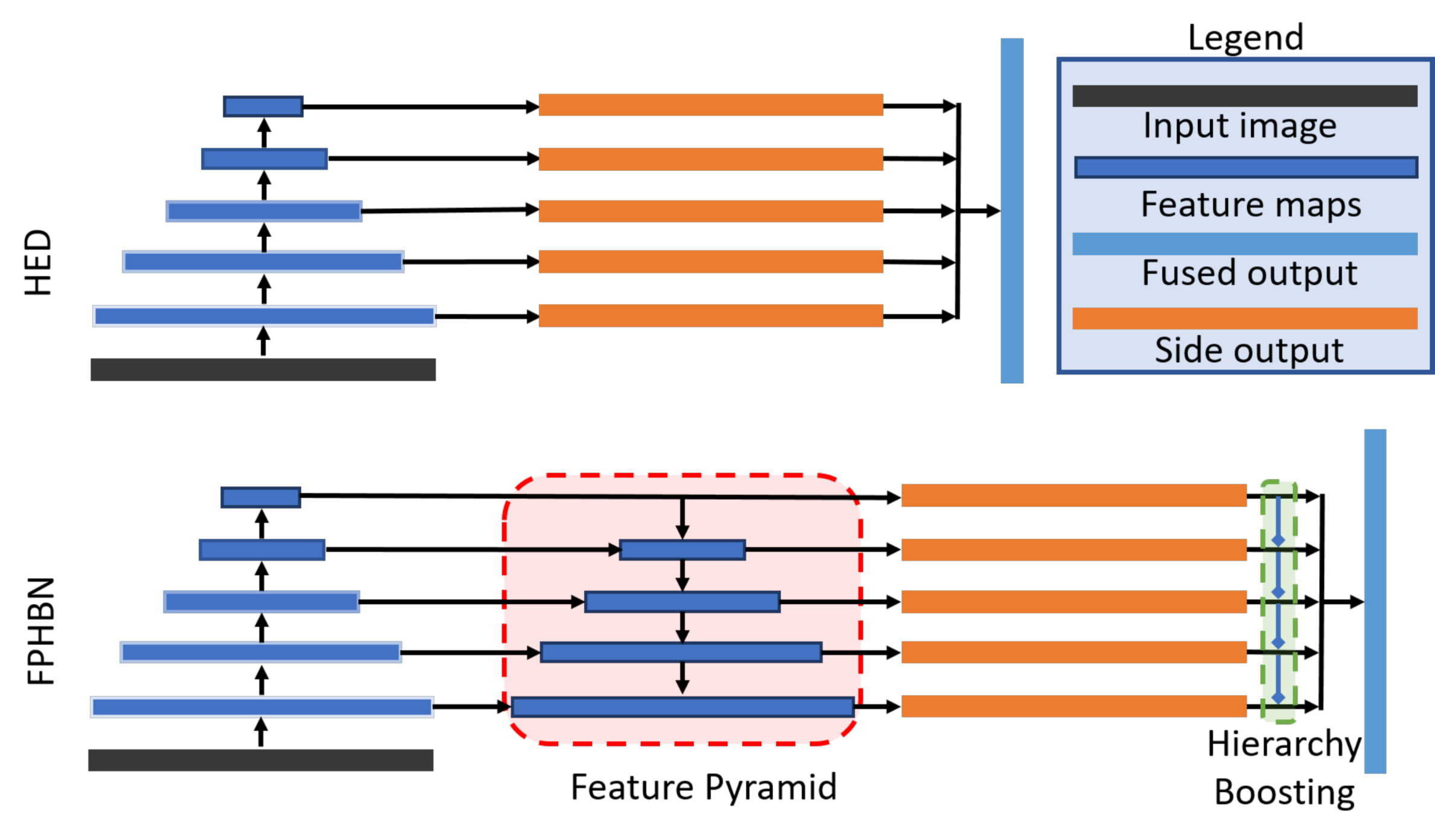}
	\caption{ The architecture of HED\cite{xie2015holistically} and FPHBN (ours). Red and green dashed boxes indicate feature pyramid and hierarchical boosting modules, respectively. The thicker outlines of feature maps means the richer context information.}
	\label{fig:hed_fphb}
\end{figure}

To overcome the aforementioned shortcomings, CrackForest \cite{shi2016automatic} incorporates complementary features from multiple levels to characterize cracks and takes advantage of the structural information in crack patches. This method is shown to outperform state-of-the-art crack detection methods like CrackTree \cite{zou2012cracktree}, CrackIT \cite{oliveira2013automatic}, Free-Form Anisotropy (FFA) \cite{nguyen2011free}, and Minimal Path Selection (MPS)\cite{amhaz2016automatic}. However, CrackForest \cite{shi2016automatic} still performs crack detection based on hand crafted feature, which is not discriminative enough to differentiate the cracks from complex background with low level cues.
\begin{figure}[!t]
	\centering
	\includegraphics[width=0.5\textwidth]{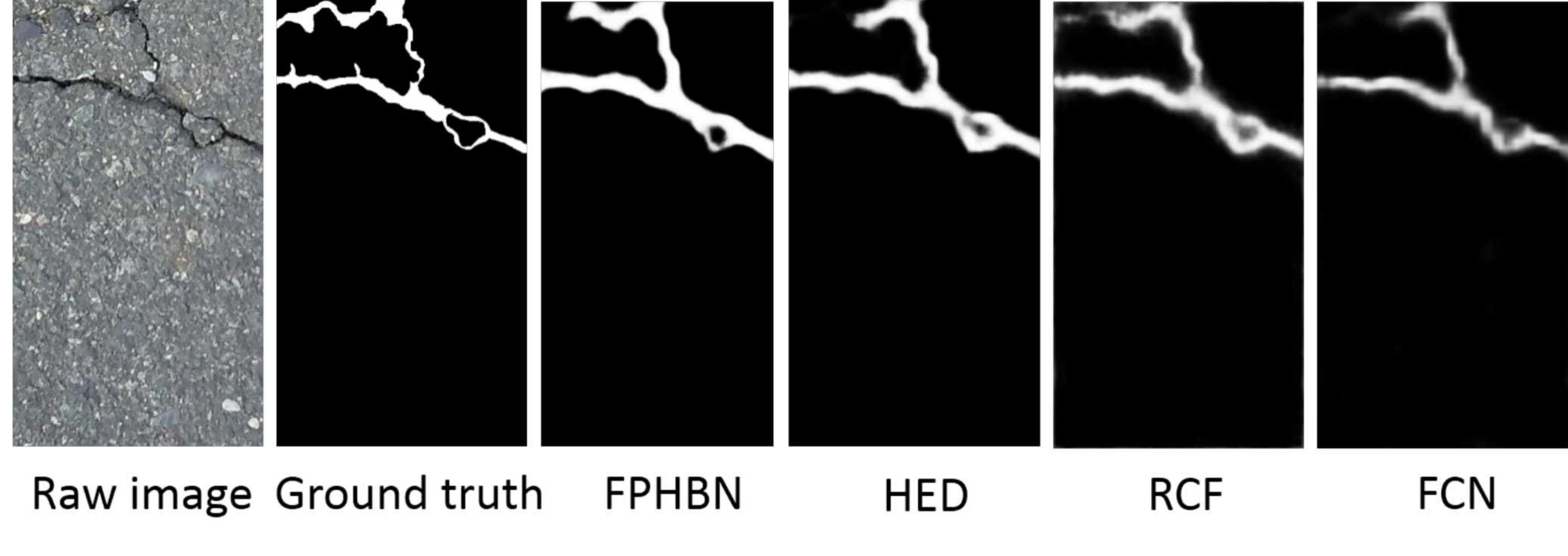}
	\caption{Visualization of crack detection results of the proposed method and state-of-the-art edge detection, semantic segmentation methods.}
	\label{fig:cmpmethod}
\end{figure}

Recently, deep learning has been widely applied in computer vision for its excellent representation capability.
Some works \cite{zhang2016road} \cite{pauly2017deeper}  \cite{eisenbach2017get} \cite{schmugge2017crack} have been devoted to leverage this property of deep learning to learn robust feature representation for crack detection. Zhang et al\cite{zhang2016road}, Pauly et al \cite{pauly2017deeper},  and Eisenbach et al \cite{eisenbach2017get} use deep learning to perform patch-based classification for crack detection,  which is inconvenient and sensitive to patch scale.  Schmugge et al \cite{schmugge2017crack} treats crack detection as a segmentation task, which classifies each pixel as crack or background category using deep learning. Although \cite{schmugge2017crack} gains decent performance, crack detection is very different from semantic segmentation in terms of ratio of foreground and background. In semantic segmentation, the foreground and background are not as unbalanced as those in crack detection.

In contrast, crack detection task is more similar to edge detection in terms of the foreground to background ratio. In addition, crack and edge detection share similar characteristics in shape and structure. Due to these common characteristics, it is intuitive to adopt edge detection methods to crack detection. For instance,
Shi et al \cite{shi2016automatic} successfully apply structure forest\cite{dollar2013structured}, a classical edge detection method, for crack detection. 
However, this approach is based on hand crafted features, lacking of representative ability.

To learn robust feature representation and to cope with the highly skewed classes problem for automatic crack detection, we propose a Feature Pyramid and Hierarchical Boosting Network (FPHBN) to automatically detect crack in an end-to-end way. FPHBN adopts Holistically-nested Edge Detection (HED) \cite{xie2015holistically}, a breakthrough edge detection method, as its backbone architecture.

Fig. \ref{fig:hed_fphb} shows the architecture configuration of HED \cite{xie2015holistically} and the proposed FPHBN.
The difference is that FPHBN integrates a feature pyramid module and a hierarchical boosting module into HED\cite{xie2015holistically}. The feature pyramid is built through a top-down architecture to introduce context information from higher-level to lower-level feature maps. This can enhance feature representation in lower-level layers, leading to a representative capability improvement to distinguish the crack from background. The hierarchical boosting is proposed to reweight samples from top layer to bottom layer, which makes the FPHBN pay more attention to hard examples. 
\begin{figure}[!t]
	\centering
	\includegraphics[width=0.5\textwidth]{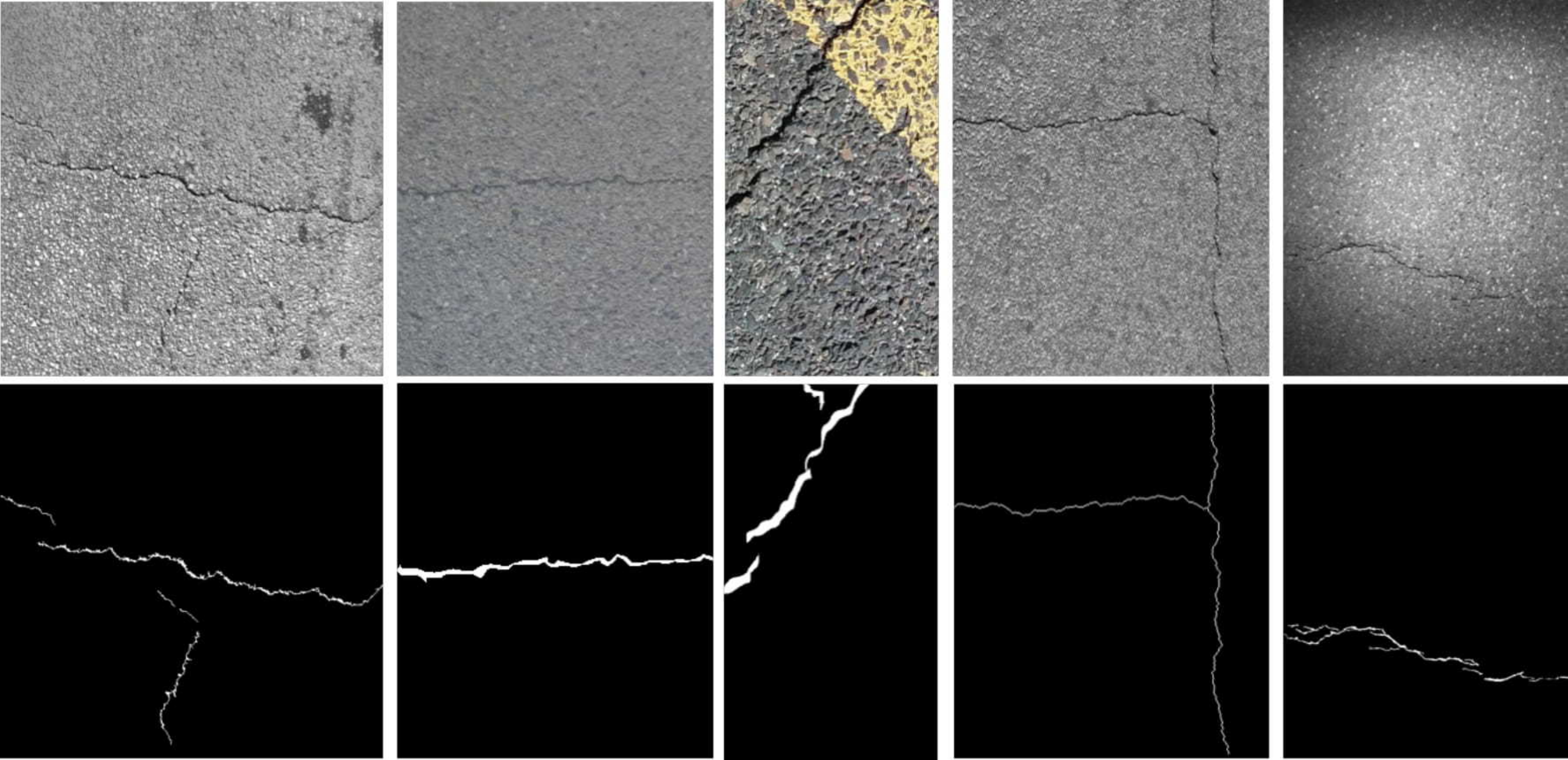}
	\caption{Representative crack images (upper row) and ground truth (lower row) from five datasets.}
	\label{fig:sampledata}
\end{figure}



Fig. \ref{fig:cmpmethod} shows the detection results of FPHBN and state-of-the-art edge and semantic segmentation
methods: HED \cite{xie2015holistically}, Richer Convolutional Features for edge detection (RCF)\cite{liu2017richer}, Fully Convolutional Networks for semantic segmentation (FCN)\cite{long2015fully}.
From Fig. \ref{fig:cmpmethod}, we can see that detection result of FPHBN is much clearer than those of other methods and has less false positives.

To evaluate crack detection algorithms, Shi et al \cite{shi2016automatic} use precision and recall (PR) as measurements. The detected pixels are treated as true positives if these pixels locate within five pixels from labeled pixels. This criterion is too loose when the crack annotation is with a large width. Moreover, PR cannot appropriately demonstrate the overlapping extent between detected crack and ground truth, especially when crack is large.
For example, in Fig. \ref{fig:sampledata}, we see that ground truth in the third column is much wider than others. In this situation, PR is not precise enough to measure the detection results.
Thus in addition to PR, we propose average intersection over union (AIU) as a complementary measurement for evaluating crack detection. By computing the average intersection over union (IU) over different thresholds, AIU takes the width information into consideration to evaluate detections and illustrates the overall overlap extent between detections and ground truth. Thus the AIU can be used to determine if the width is precisely estimated, which is critical to assess the damage degree of pavement.

The contributions of this paper can be summarized in the following aspects:
\begin{itemize}
\item  A feature pyramid module is introduced for crack detection. The feature pyramid is constructed by a top-down architecture which incorporates context information from top to bottom, layer by layer.

\item A hierarchical boosting module is proposed to reweight samples layer by layer so as to make the model focus on hard samples.
\item{A new measurement is proposed to evaluate crack detection methods. The measurement not only takes into account the crack width, but also avoids the annotation bias.}
\end{itemize}

The rest of the paper is organized in the following way: Related works are reviewed in Section \ref{related works}; Section \ref{method} describes the details of the proposed FPHBN; Section \ref{Experiments and Results} demonstrates experiments design and analyzes experimental results. Section \ref{conclusion} is the conclusion.




\section{Related works}
\label{related works}
In this section, we first briefly review traditional works on crack detection. Then, deep learning-based crack detection methods are discussed to demonstrate their superiority to traditional approaches.


\subsection{Traditional crack detection methods}
In this work, we refer to as traditional crack detection methods the crack detection methods that are based on non-deep learning techniques. Over the past years, numerous researchers have been devoted to automating crack detection. These works can be divided into five categories: 1) wavelet transform, 2) image thresholding, 3) hand crafted feature and classification, 4) edge detection-based methods, and 5) minimal path-based methods.
\subsubsection{Wavelet transform}
\begin{figure*}[!t]
	\centering
	\includegraphics[width=1\textwidth]{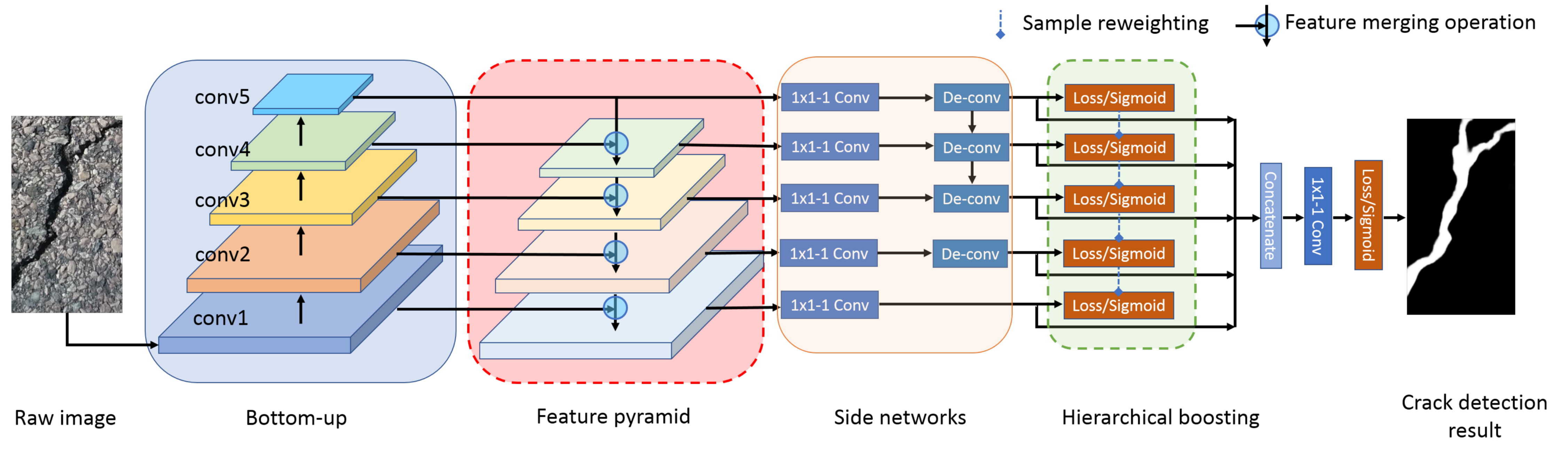}
	\caption{The proposed network architecture consists of four components. Bottom-up is to construct multi-scale features; Feature pyramid is used to introduce context information to lower-level layers; Side networks are to carry out deep supervised learning; Hierarchical boosting is to down-weight easy samples and up-weight hard samples.}
	\label{fig:architecture}
\end{figure*}
In \cite{zhou2006wavelet}, a wavelet transform is applied to a pavement image, such that the image is decomposed into different frequency subbands. The distresses and noise are transformed into high and low amplitude wavelet coefficients, respectively. Subirats et al \cite{subirats2006automation} build a complex coefficient map by performing a 2D wavelet transform in a multi-scale way; then crack region can be obtained by searching the maximum wavelet coefficients from largest to smallest scale. However, these approaches cannot deal with cracks with low continuity or high curvature property since the anisotropic characteristic of wavelet.
\subsubsection{Image thresholding}
In \cite{huang2012novel,peng2015research,xu2013pavement,chambon2011automatic}, preprocessing algorithms are first used  to reduce the illumination artifacts. Then thresholding is applied to the image to yield crack candidates. The processed crack image is further refined using morphological technologies. \cite{zou2012cracktree} \cite{fernandes2014pavement} \cite{tang2013automatic} are variants of this group, which leverage graph-based methods for crack candidates refinement.
\subsubsection{Hand crafted feature and classification}
Most of the current crack detection methods are based on hand crafted features and patch-based classifier. In \cite{kapela2015asphalt,quintana2016simplified,varadharajan2014vision,zakeri2013multi,hu2010novel}, hand crafted features, e.g., HOG\cite{kapela2015asphalt}, LBP\cite{hu2010novel}, are extracted  from an image patch as a descriptor of crack, followed by a classifier, e.g., support vector machine. 
\subsubsection{Edge detection-based methods}
 Yan et al \cite{maode2007pavement} introduce morphological filters into crack detection and removes noise with a modified median filter. Ayenu-Prah et al \cite{ayenu2008evaluating} apply Sobel edge detector to detect crack after smoothing image and removing speckle noise by a bidimensional empirical mode decomposition algorithm.  Shi et al \cite{shi2016automatic} apply random structure forest \cite{dollar2013structured} to exploit structural information for crack detection.
\subsubsection{Minimal path-based methods}
Kass et al \cite{kass1988snakes} propose to use minimal path method to extract simple open curves in images for given both endpoints of a curve. Kaul et al\cite{kaul2012detecting} propose to detect same type of contour-like image structures using an improved minimal path method. The improved method needs less prior knowledge of both topology and endpoints of the desired curves. Amhaz et al\cite{amhaz2016automatic} propose a two-stage method for crack detection: first select endpoint at local scale; second select minimal paths at global scale. Nguyen et al\cite{nguyen2011free} present a method to take into account intensity and crack shape features for crack detection simultaneously by introducing Free-Form Anisotropy\cite{nguyen2011free}.
\subsection{Deep learning-based crack detection}

Recent years, deep learning achieves unprecedented success in computer vision \cite{krizhevsky2012imagenet}. A lot of works try to apply deep learning to crack detection task. Zhang et al \cite{zhang2016road} first propose a relatively shallow neural network, consisting of four convolutional layers and two fully connected layers, to perform crack detection in a patch-based way. Moreover, Zhang et al \cite{zhang2016road} compare their method with hand crafted feature based methods to demonstrate the advantages of feature representation of deep learning.
Pauly et al\cite{pauly2017deeper} apply a deeper neural network to classify crack and non-crack patches and demonstrates the superiority of deeper neural network. Feng et al\cite{feng2017deep} propose a deep active learning system to deal with limited label resources problem. Eisenbach et al \cite{eisenbach2017get} present a road distress dataset for training deep learning network and first evaluates and analyzes state-of-the-art approaches in pavement distress detection.

The aforementioned approaches treat crack detection as a patch-based classification task. Each image is cropped to small patches, then a deep neural network is trained to classify each patch as crack or not. This way is inconvenient and sensitive to patch scale.
Due to the rapid development in semantic segmentation task \cite{long2015fully} \cite{badrinarayanan2015segnet}\cite{fan2018multi}, Schmugge et al \cite{schmugge2017crack} present a crack segmentation method based on SegNet \cite{badrinarayanan2015segnet} for remote visual examination videos.
This method performs crack detection by aggregating the crack probability from multiple overlapping frames in a video.

\section{method}
\label{method}
\subsection{Overview of proposed method}
In this paper, crack detection is formulated as a pixel-wise binary classification task. For a given crack image, a designed model yields a crack prediction map, where crack regions have higher probability and non-crack regions have lower probability.
Fig. \ref{fig:architecture} shows the architecture of the proposed Feature Pyramid and Hierarchical Boosting Network (FPHBN). FPHBN is composed of four major components: 1. a bottom-up architecture for hierarchical feature extraction, 2. a feature pyramid for merging context information to lower layers using a top-down architecture, 3. side networks for deep supervision learning, and 4. a hierarchical boosting module to adjust sample weights in a nested way.
\begin{figure}[!t]
	\centering
	\includegraphics[width=0.5\textwidth]{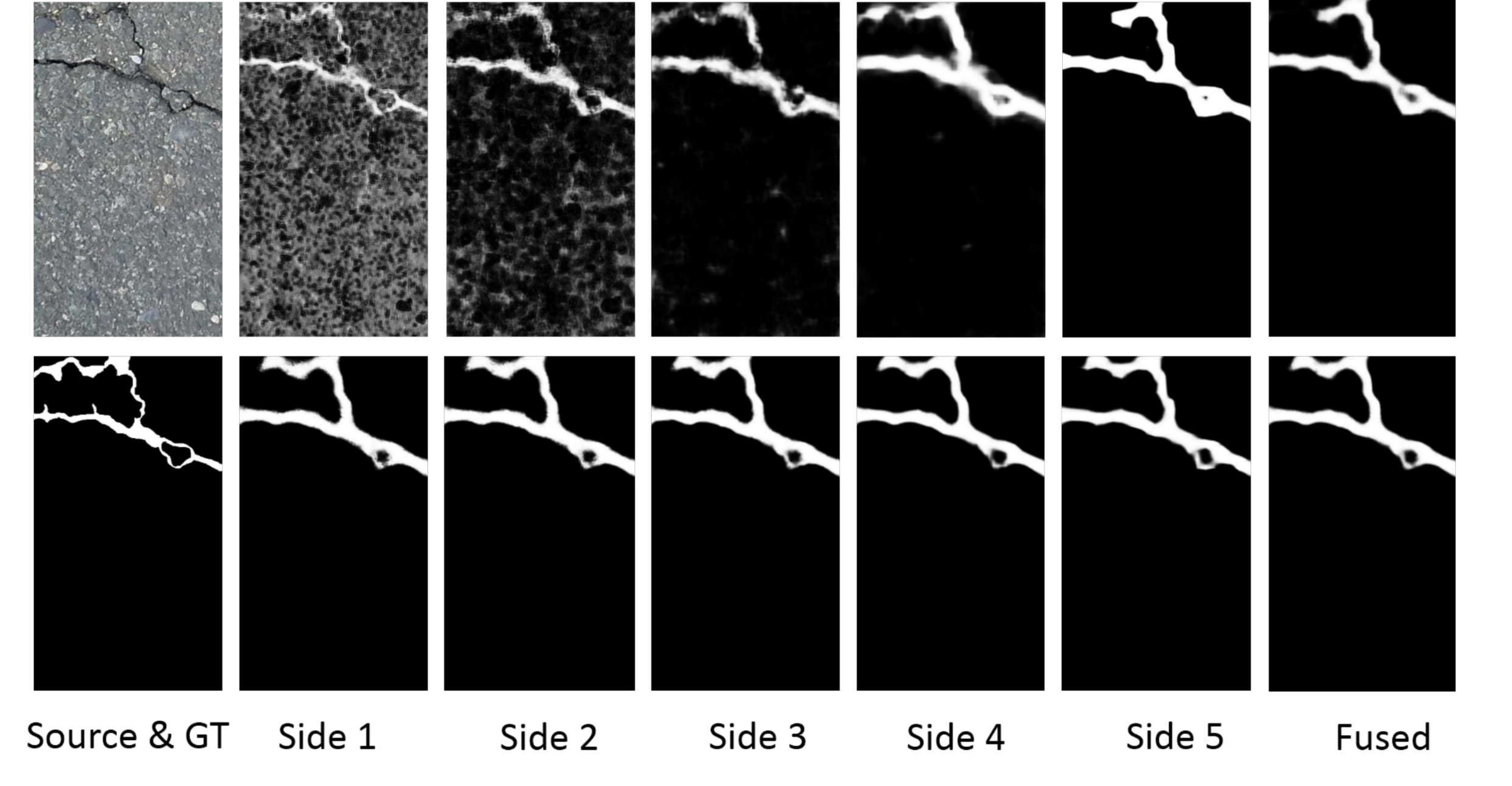}
	\caption{Visual comparison of crack detection of HED\cite{xie2015holistically} and HED\cite{xie2015holistically} with feature pyramid (HED-FP). Top row is the results of HED \cite{xie2015holistically}; bottom is the results of HED-FP.}
	\label{fig:cmp}
\end{figure}

Given a crack image and corresponding ground truth, the crack image is first fed into the bottom-up network to extract features of different levels.
Each conv layer corresponds to a level in the pyramid. At each level, except for the fifth one, a feature merging operation is conducted to incorporate higher-level feature maps to lower-level ones layer-by-layer to make context information flow from higher to lower ones. At each level the feature maps in top-down architecture are fed to a convolutional filter of size $1 \times 1$ for dimension reduction and a de-convolutional filter to resize feature map to the same size of the input image. Then each resized feature map is introduced to the hierarchical boosting module to yield crack prediction map and compute sigmoid cross-entropy loss with ground truth.
The convolutional filter, de-convolutional filter, and loss layer at each level comprise a side network.
Finally, all the five resized feature maps are fused together by a concatenate layer followed by a $1 \times 1$ convolutional filter to produce a crack prediction map and compute a final sigmoid cross-entropy loss with ground truth.
\subsection{Bottom-up architecture}
In our method, the bottom-up architecture consists of the conv1-conv5 parts of VGG \cite{simonyan2014very} including max pooling layers between convolutional layers. Through the bottom-up architecture, convolutional networks (ConvNets) compute a hierarchical feature representation.  Due to the max pooling layers, the feature hierarchy has an inherent multi-scale, pyramid shape.

Unlike multi-scale features built upon multi-scale images, the multi-scale feature computed by deep ConvNets is more robust to variance in scale\cite{lin2016feature}. Thus it facilitates recognition on a single input scale. More importantly, deep ConvNets make the progress of producing multi-scale representation automated and convenient. This is much more efficient and effective than computing engineered features on various scales of images.

In the bottom-up network, the in-network hierarchy produces feature maps of different spatial resolutions, but introduces large context gaps. 
The bottom levels of feature maps are of higher resolution but have less context information than top levels of features.
In contrast, the top levels of feature maps are lower resolution but have more context information. The context information is helpful for crack detection. Therefore, if these feature maps are directly fed into each side network, the output from different side network varies significantly.

The first row in Fig. \ref{fig:cmp} shows the five side outputs and fused output. We note that the side outputs 1-3 are very messy and hardly recognized; the side outputs 4-5 and fused are much better than the side outputs 1-3. In addition, we find that although fused result is clearer than side outputs 1-4, it is still a little blur. This is because fused result merges the information from the side outputs 1-4, which is cluttered and blur. The reason causes these messy side outputs is that the lower-level layers lack context information.
\subsection{Feature pyramid}
\begin{figure}[!t]
	\centering
	\includegraphics[width=0.5\textwidth]{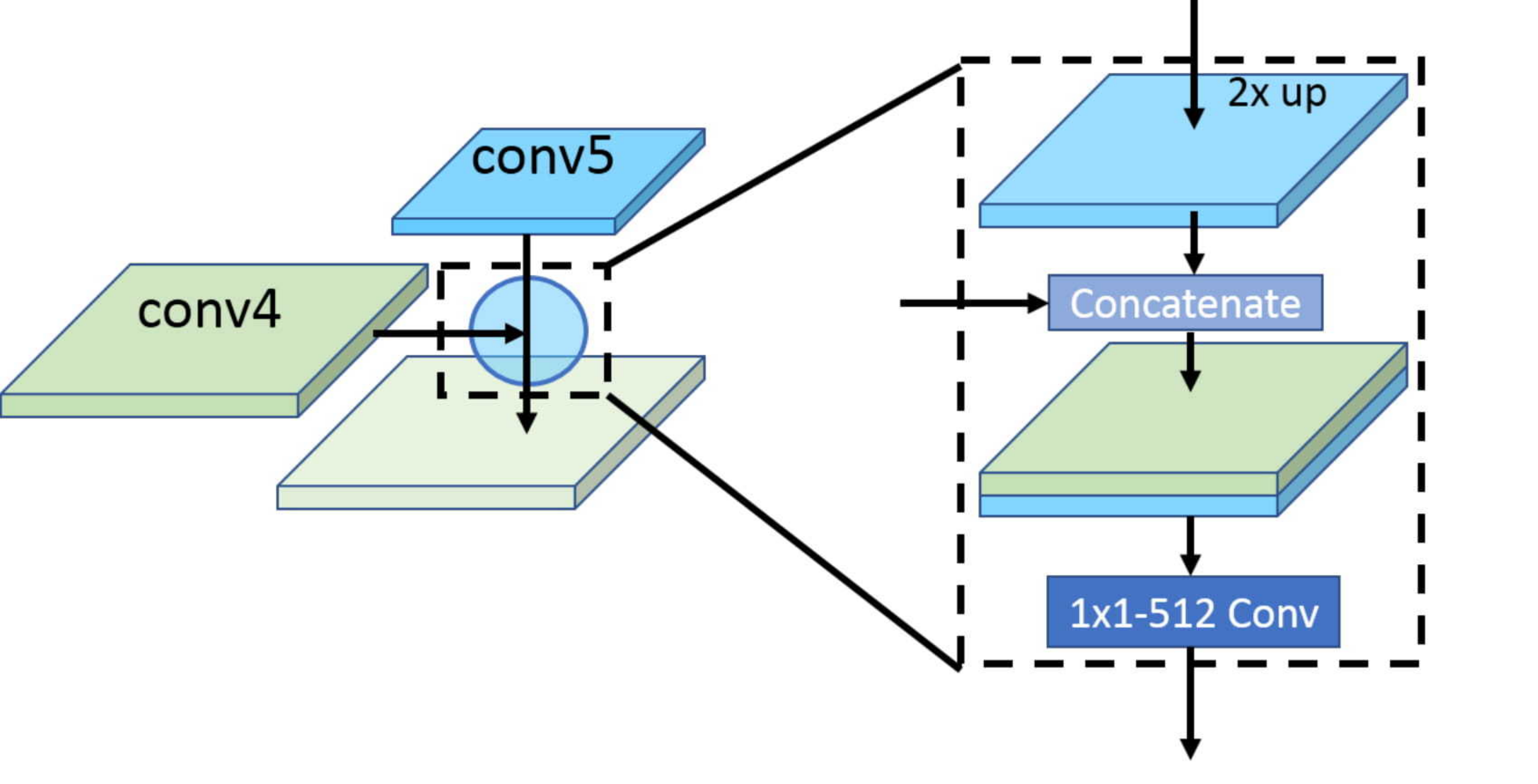}

	\caption{Illustration of feature merging operation in feature pyramid. The feature maps from conv5 are upsampled twice first and then concatenated with the feature maps from conv4.}
	
	\label{fig:merge}
\end{figure}

To deal with the aforementioned issue, we introduce context information to lower-level layers to generate a feature pyramid through a top-down architecture, which is inspired by \cite{lin2016feature}.
As shown in Fig. \ref{fig:architecture}, the top-down architecture feeds the fifth and fourth levels of feature maps into a feature merging operation unit, then feeds the output and third level of feature maps to next merging operation unit.  This operation is conducted progressively until the bottom level. In this way, we produce a set of new feature maps, which contain much richer features at each level than that in bottom-up network except for fifth level.
From Fig. \ref{fig:cmp}, we note that after introducing feature pyramid, the side outputs 1-4 and fused result are significantly clearer than counterparts from HED \cite{xie2015holistically}. Combining multi-scale context information into low-level feature maps by the feature pyramid module benefits the performance improvement of low-level side networks.
Specifically, detecting various scales of crack requires different extents of context information.
Since the deeper convolution layers have much larger
receptive fields than the shallower ones, the high-level
features contain more context information than the low-level
ones. Therefore, when high-level features are combined with low-level ones, the low-level side networks utilize
the multi-scale context information to increase its detection
performance and to boost the fused detection performance.

The feature merging operation of the feature pyramid is illustrated in Fig. \ref{fig:merge}. We use the feature merging operation at the fourth level as an example to demonstrate the concrete operations. The feature maps from conv5 are upsampled twice and concatenated with the feature maps from conv4. The concatenated feature maps are fused and reduced by a $1\times1$ convolutional layer of 512 filters. Note that from conv3 to conv1 level, the filter number is set to 256, 128, and 64, respectively.

\subsection{Side networks}
The side network at each level performs crack prediction individually. 
This configuration of learning is named as deeply supervised learning, proposed in \cite{lee2015deeply}  and justified considerably useful for edge detection in HED \cite{xie2015holistically}.
The key characteristic of the deep supervision is that instead of performing recognition task at the highest layer, the recognition is conducted at each level through the side network.
Let us first introduce HED \cite{xie2015holistically} in the context of crack detection.

\subsubsection{Training Phase}

 We denote our crack training dataset as $S={(X_n,Y_n),n=1,...,N}$, where sample $X_n$ and $Y_n$ denote the raw crack image and the corresponding binary ground truth crack map, respectively. 
 For convenience, we drop the subscript $n$ in subsequent paragraphs. 
 The parameters of the entire network are denoted as $W$. Assume there are $M$ side networks in HED\cite{xie2015holistically}. Each side network is associated with a classifier, in which the corresponding weights are denoted as $w=(w^{(1)},..., w^{(M)})$. The objective function is defined as
 \begin{equation} \label{eq1}
 \begin{split}
 \mathcal{L}_{side}(W,w)=\sum_{m=1}^{M} \ell_{side} ^{m}(W,w^{(m)}),
 \end{split}
 \end{equation}
 where $\ell_{side}$ represents the image-level loss function for a side network. During the image-to-image training, the loss function is computed over all pixels in a training image $X=(x_i,i=1,...,|X|)$ and crack map $Y=(y_i, i=1,..., |Y|), y_i \in \{0,1\}$. However, this kind of loss function regards positives and negatives equally, which is not suited for practical situation. For a typical crack image, as shown in Fig. \ref{fig:cmp}, the distribution of crack and non-crack pixels is heavily biased. Most regions of the image are non-crack pixels. 
 HED\cite{xie2015holistically} uses a simple strategy to automatically balance the contribution to the loss from positives and negatives. A class-balancing weight $\beta$ is introduced in a pixel-wise way. Index $i$ is over the image spatial dimensions of image $X$. Then $\beta$ is used to offset the imbalance between crack and non-crack pixels. Specifically, Equation \ref{eq1} is rewritten as
 \begin{equation} \label{eq2}
 \begin{aligned}
\ell_{side}^{(m)}(W,w^{(m)})=-\beta \sum_{i \in Y_{+}}\log{P_i(y_i=1|X;W,w^{(m)})}\\
 -(1-\beta)\sum_{i\in Y_{-}}\log{P_i(y_i=0|X;W,w^{(m)})},
 \end{aligned}
 \end{equation}
where $\beta =|Y_{-}|/|Y|$ and $1-\beta=|Y_{+}|/|Y|$. $|Y_{-}|$ and $|Y_{+}|$ denote the crack and non-crack pixels in ground truth image, respectively. $ P_i(y_i=1|X;W,w^{(m)})=\sigma (a_{i}^{m}) \in [0,1]$ is computed by a sigmoid function $\sigma (.)$ on the activation at pixel $i$. At each side output network, we then obtain crack map predictions $\hat{Y}_{side}^{(m)}=\sigma(\hat{A}_{side}^{(m)})$, where $\hat{A}_{side}^{(m)}\equiv \{a_i^{(m)},i=1,...,|Y|\}$ are activations of the side output of layer $m$.

To leverage the side output prediction, a `weighted-fusion' layer is added to the network and simultaneously learns the fusion weight with all side networks during training. The loss function of the fusion layer $\mathcal{L}_{fuse}$ becomes
 \begin{equation} \label{eq3}
\mathcal{L}_{fuse}(W,w,h)=\mathrm{Ds}(Y,\hat{Y}_{fuse}),
 \end{equation}
 where $\hat{Y}_{fuse}\equiv \delta(\sum_{m=1}^{M}h_m\hat{A}_{side}^{(m)})$ where $h=(h_1,...,h_M)$ is the fusion weight. $Ds(.,.)$ is the distance between the fused predictions and the ground truth crack map, which is set to a sigmoid cross-entropy loss. For the entire network, the overall object function needs to be minimized is
   \begin{equation} \label{eq4}
  (W,w,h)^*=\arg\min(\mathcal{L}_{side}(W,w)+\mathcal{L}_{fuse}(W,w,h)), \end{equation}

\subsubsection{Test Phase}
 During testing, for image $X$, a set of crack map predictions are yielded from both side network and the fusion layer:
 \begin{equation} \label{eq5}
 (\hat{Y}_{fuse},\hat{Y}_{side}^{(1)},...,\hat{Y}_{side}^{(M)})=\mathrm{CNN}(X,(W,w,h)^*),
 \end{equation}
 where CNN(.) denotes the crack maps generated by HED\cite{xie2015holistically}.

\subsection{Hierarchical boosting}
Although HED \cite{xie2015holistically} takes into account the issue of unbalanced positives and negatives by using a class-balancing weight $\beta$, it cannot differentiate easy and hard samples. Specifically, in crack detection, the loss function Equation \ref{eq4} is dominated by easily classified negative samples since the samples are highly skewed. Thus, the network cannot effectively learn parameters from misclassified samples during training phase.

To address this problem, a common solution is to perform some forms of hard mining \cite{shrivastava2016training}\cite{viola2001rapid} that samples hard samples during training or more complex sampling/reweighting schemes \cite{bulo2017loss}. In contrast, Lin et al \cite{lin2017focal} propose a novel loss to down-weight well-classified examples and focus on hard examples during training.

Different from these previous works, we design a new scheme, hierarchical boosting, to reweight samples. 
From Fig. \ref{fig:architecture}, we note that in the hierarchical boosting module, from top to bottom, a sample reweighting operation is conducted layer by layer. This top-down hierarchical is inspired by the observation that in Fig. \ref{fig:cmp} the side output predictions are similar with each other. This means each side network has similar capability for crack detection.
In addition, the five side networks are sequential in terms of information flowing. For example, the fourth side network needs the feature maps from the fifth side network to compute crack prediction map  and loss.
If the upper network can inform lower network which samples are hard to classify, the lower networks will pay more attention to these hard samples. Through this communication, crack detection performance can be improved. Therefore, we propose the hierarchical boosting method to deal with the hard sample problem by facilitating communication between adjacent side networks.

Specifically, given $P^{m+1}$ the output from the $m+1$ th side network, the difference between $P^{m+1}$ and ground truth is denoted as $D^{m+1}=(d^{m+1}_i,i=1,...,|Y|)$. Thus, the loss function Equation \ref{eq2} of the $m$-th side network can be rewritten to
 \begin{equation} \label{eq6}
 \begin{aligned}
 \ell_{side}^{(m)}(W,w^{(m)})=-\beta \sum_{i \in Y_{+}}|d_i^{m+1}|\log{P_i(y_i=1|X;W,w^{(m)})}\\
 -(1-\beta)\sum_{i\in Y_{-}}|d_i^{m+1}|\log{P_i(y_i=0|X;W,w^{(m)})},
 \end{aligned}
 \end{equation}
 where $d_i^{m+1}$ is the difference at pixel $i$, $m$ is integer and $1\leq m\leq M-1$.
 Thus, the prediction of deeper side network is leveraged to weight loss term in the shallower side network to encourage the communication among different side networks.
 \section{Experiments and Results}
 \label{Experiments and Results}
 In this section, we describe the implementation details of the proposed FPHBN. Then datasets for evaluation, compared methods, and evaluation criteria are introduced. Finally, we present and analyze experimental results.
 \subsection{Implementation details}
 The proposed method is implemented on the widely used Caffe library \cite{jia2014caffe} and an open implementation of FCN\cite{long2015fully}. The bottom-up part is the conv1-conv5 of pretrained VGG \cite{simonyan2014very}. The feature pyramid is implemented by using Concat and Convolutional layers in Caffe. Sample reweighting is implemented using python.
 \subsubsection{Parameters setting}
  The hyperparameters include: mini-batch (10), learning rate (1e-8), loss weight of each side network (1), momentum (0.9), weight decay (0.0002), initialization of the each network (0), initialization of fusion layer (0.2), initialization of filters in feature merging operation (Gaussian kernel with mean 0 and std 0.01), the number of training iteration (40,000), learning rate divided by 10 per 10,000 iterations. The model is saved every 4,000 iterations.
 \subsubsection{Upsampling operation}
  Within the proposed FPHBN, the upsampling operations are implemented with in-network de-convolutional layers. Instead of learning the parameters of de-convolutional layers, we freeze the parameters to perform bilinear interpolation.
 \subsubsection{Sample reweighting}
 For sample reweighting, as there is no layer in Caffe to complete such function, we implement it with a python layer and integrate it to the proposed network using python Caffe interface. To make sure the performance gained is not caused by our implementation, we first conduct experiments using our implemented class-imbalance cross-entropy loss and compare it with the original implementation. We find that the experimental results are same.

 \subsubsection{Computation platform}
 At inference phase, deep learning based methods are tested on a 12G GeForce GTX TITAN X. Non-deep learning method is tested on a computer with 16G RAM and i7-3770 CPU@3.14GHz.
 \subsection{Datasets}
 \subsubsection{CRACK500}
 We collect a pavement crack dataset with 500 images of size around $2,000 \times 1,500$ pixels on main campus of Temple University using cell phones. This dataset is named as CRACK500.
 Each crack image has a pixel-level annotated binary map. To facilitate future research, we share the CRACK500 to the research community, To our best knowledge, this dataset is currently the largest publicly accessible pavement crack dataset with pixel-wise annotation. The dataset is divided into 250 images of training data, 50 images of validation data, and 200 images of test data.

 Due to limited number of images, large size of each image, and restricted computation resource, we crop each image into 16 non-overlapped image regions and only the region containing more than 1,000 pixels of crack is kept. Through this way, the training data consists of 1,896 images, validation data contains 348 images, test data contains 1124 images. 
 The validation data is used to choose the best model during training process to prevent overfitting. Once the model is chosen, it is tested on the test data and other datasets for generalizability evaluation.
 \subsubsection{GAPs384}
 German Asphalt Pavement Distress (GAPs) dataset is presented in \cite{eisenbach2017get} to address the issue of comparability in the pavement distress domain by providing a standardized high-quality dataset of large scale. The GAPs dataset includes a total of 1,969 gray valued images, with various classes of distress such as cracks, potholes, inlaid patches, et. al. The image resolution is $1,920 \times 1,080$ pixels with a per pixel resotution of $1.2mm \times 1.2mm$. For more details of the dataset, the readers are referred to \cite{eisenbach2017get}

 The actual damage in an image is enclosed by a bounding box. This type of annotation is not fine enough to training deep model for a pixel-wise crack prediction task. To address this problem, we manually select 384 images from the GAPs dataset, which only includes crack class of distress, and conduct pixel-wise annotation. This pixel-wise annotated crack dataset is named as GAPs384 and used to test the generalization of the model trained on CRACK500.

 Due to the large size of image and limited memory of GPU, each image is cropped to 6 non-overlapped image regions of size $640\times540$ pixels. Only the image regions with more than $1,000$ pixels are remained. Thus we gain 509 images for test. 
 \subsubsection{Cracktree200}
Zou et al\cite{zou2012cracktree} present a dataset to evaluate their proposed method. The dataset includes 206 pavement images of size $800\times 600$ with various types of cracks. Therefore, we name this dataset as Cracktree200. This dataset is with challenges like shadows, occlusions, low contrast, noise, etc. The annotation of the dataset is pixel-wise label, which can be directly used for evaluation.

\subsubsection{CFD}
Shi et al\cite{shi2016automatic} propose an annotated road crack dataset called CFD. The dataset consists of 118 images of size $480 \times 320$ pixels. Each image has manually labeled crack contours. The device used to acquire the images is an iPhone5 with focus of 4mm, aperture of f/2.4 and exposure time of 1/135s. The CFD is used for evaluating model.
\subsubsection{Aigle-RN \& ESAR \& LCMS}
Aigle-RN is proposed in \cite{amhaz2016automatic}, which contains 38 images with pixel-level annotations. The dataset is acquired at traffic speed for periodically monitoring the French pavement surface condition using Aigle-RN system. ESAR is acquired by a static acquisition system with no controlled lighting. ESAR has 15 fully annotated crack images. LCMS contains 5 pixel-wise annotated crack images.
Since the three datasets have small number of images, they are combined to one dataset named AEL for model evaluation.
\subsection{Compared methods}
\subsubsection{HED}
HED \cite{xie2015holistically} is a breakthrough work in edge detection. We train HED\cite{xie2015holistically} for crack detection on Crack500 training data and choose the best model using Crack500 validation data. During training the hyperparameters are set as in FPHBN except for the feature merging operation unit.
\subsubsection{RCF}
RCF \cite{liu2017richer} is an extension work based on HED \cite{xie2015holistically} for edge detection. The training and validation procedure are same as in HED\cite{xie2015holistically}. The hyperparameters are set same as those in HED \cite{xie2015holistically} except the learning rate, which is set to 1e-9.
\subsubsection{FCN}
We adopt FCN-8s \cite{long2015fully} by replacing the loss function with sigmoid cross-entropy loss for crack detection. The training and validation data used are same with those in HED\cite{xie2015holistically}. The hyperparameters: base learning rate is set to 0.00001, momentum is set to 0.99, weight decay is set to 0.0005.

\subsubsection{CrackForest}
 We train CrackForest\cite{shi2016automatic} on CRACK500 training data. All the hyperparameters are set as default.
   \begin{figure}[!t]
   	\centering
   	\includegraphics[width=0.5\textwidth]{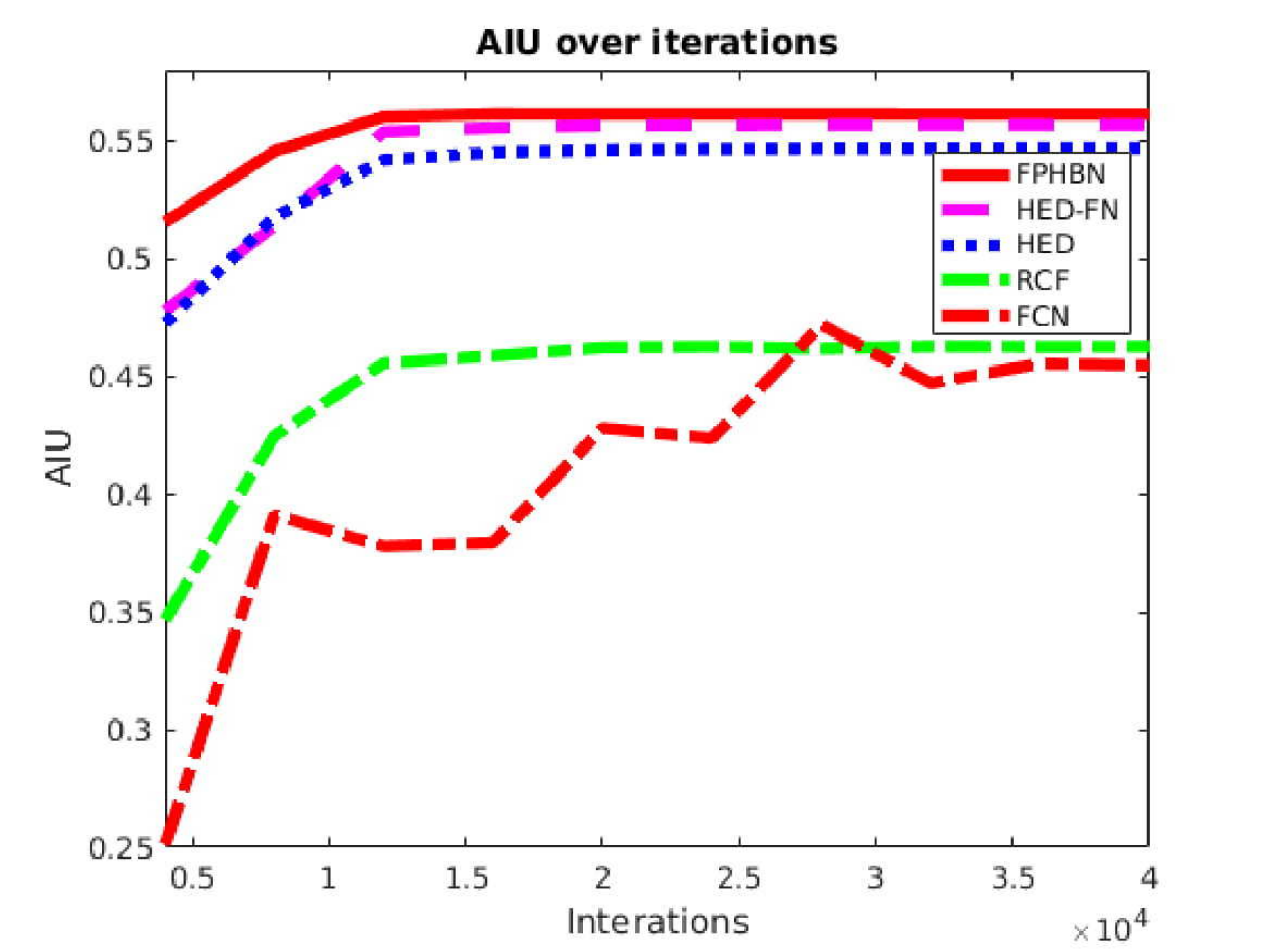}
   	\caption{The curves of AIU over training iterations of FPHBN, HED-FP, HED\cite{xie2015holistically}, RCF\cite{liu2017richer}, FCN\cite{long2015fully} on CRACK500 validation data.}
   	\label{fig:AIUval}
   \end{figure}
   	
   	

    \begin{table}[t]
    	\centering
    	\small

    	\caption{The AIU of  FPHBN, HED-FP, HED\cite{xie2015holistically}, RCF\cite{liu2017richer}, FCN\cite{long2015fully} at the chosen iteration on the CRACK500 validation dataset.}	
    	    	\vspace{-0.1in}
    	\begin{tabular}{c|c|c|c|c|c}
    		\hline\hline
    		Methods& \textbf{FPHBN} &HED-FP& HED& RCF&FCN\\
    		\hline
    		AIU &\textbf{0.560} & 0.553 &0.541 &0.455&0.455\\ \hline
    	\end{tabular}

    	\label{table:crack500val}

    \end{table}
\subsection{Evaluation criteria}
 Since the similarity with edge detection, it is intuitive to directly leverage criteria of edge detection to conduct evaluation for crack detection. The standard criteria in edge detection domain are the best F-measure on the data set for a
 fixed scale (ODS), the aggregate F-measure on the data set
 for the best scale in each image (OIS).

 The definitions of the ODS and OIS are $\max\{2\frac{P_t\times R_t}{P_t+R_t}: t=0.01,0.02,...,0.99
 \}$ and $\frac{1}{N_{img}}\sum_{i}^{N_{img}}\max\{2\frac{P_t^i\times R_t^i}{P_t^i+R_t^i}: t=0.01,0.02,...,0.99
 \}$. The $t$ denotes the threshold, $i$ is the index of image, $N_{img}$ is the total number of images. $P_t$ and $R_t$ are precision and recall at threshold $t$ over dataset. $P_t^i$ and $R_t^i$ are computed over image $i$.
 For the detailed definition of the two criteria, the readers are referred to \cite{arbelaez2011contour}.
   The edge ground truth annotation is a binary boundary map, which is different from the ground truth annotation in some crack datasets, where crack annotation is a binary segmentation map. Therefore, during evaluation both crack detection and ground truth are first processed by non-max suppression (NMS), then thinned to one pixel wide before computing ODS and OIS. Note that the maximum tolerance allowed for correct matches of prediction and ground truth is set to 0.0075.

  The proposed new measurement, AIU, is computed on the detection and ground truth without NMS and thinning operation.
    AIU of an image is defined as
    	\vspace{-0.1in}
\begin{equation} \label{eq7}
 \frac{1}{N_t}\sum_{t} \frac{N_{pg}^t}{N_p^t+N_g^t-N_{pg}^t}
\end{equation}
where $N_t$ denotes the total number of thresholds $t \in [0.01,0.99]$ with interval $0.01$; for a given threshold t, $N_{pg}^t$ is the number of pixels of intersected region between the predicted and ground truth crack area; $N_p^t$ and $N_g^t$ denote the number of pixels of predicted and ground truth crack region, respectively. Thus the AIU is in the range of 0 to 1, the higher value means the better performance. The AIU of a dataset is the average of the AIU of all images in the dataset.
\begin{figure*}[!t]
	\centering
	\includegraphics[width=1\textwidth,height=.35\linewidth]{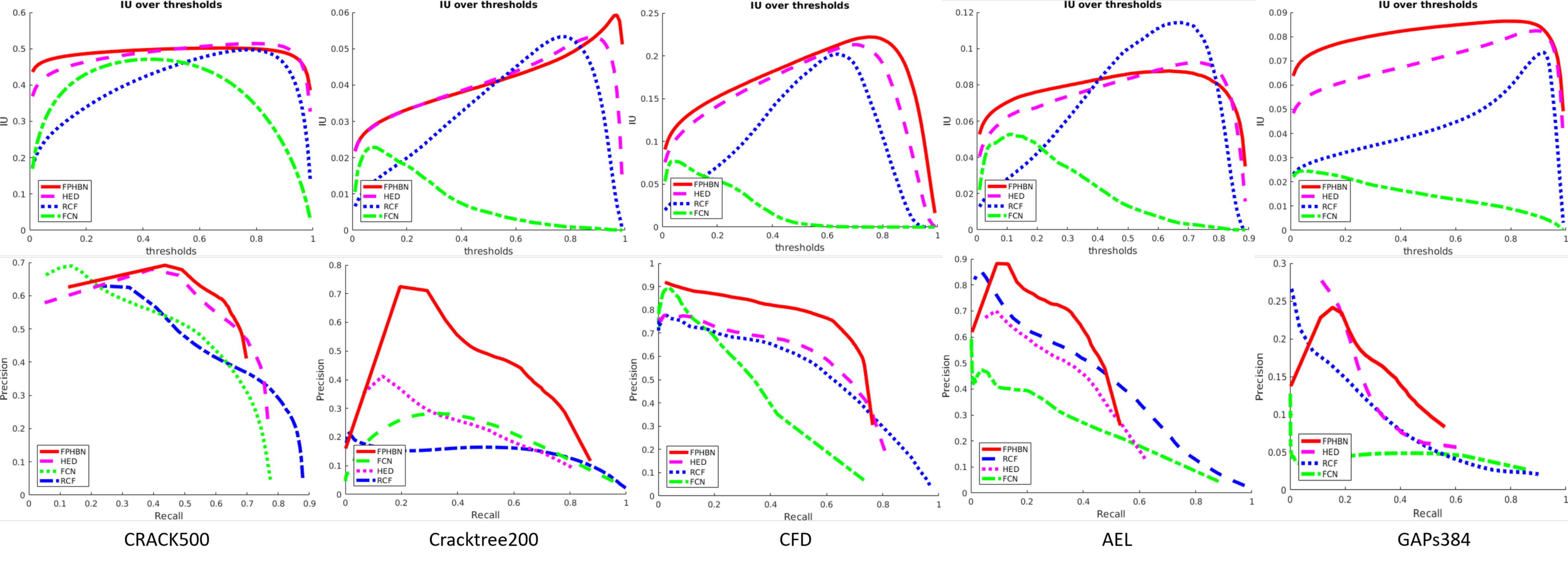}
	\caption{The evaluation curve of compared methods on five datasets. The upper row is curve of IU over thresholds. The lower row is PR curve.}
	\label{fig:curve}
\end{figure*}
\begin{figure*}[!t]
	\centering
	\includegraphics[width=1\textwidth,height=.4\linewidth]{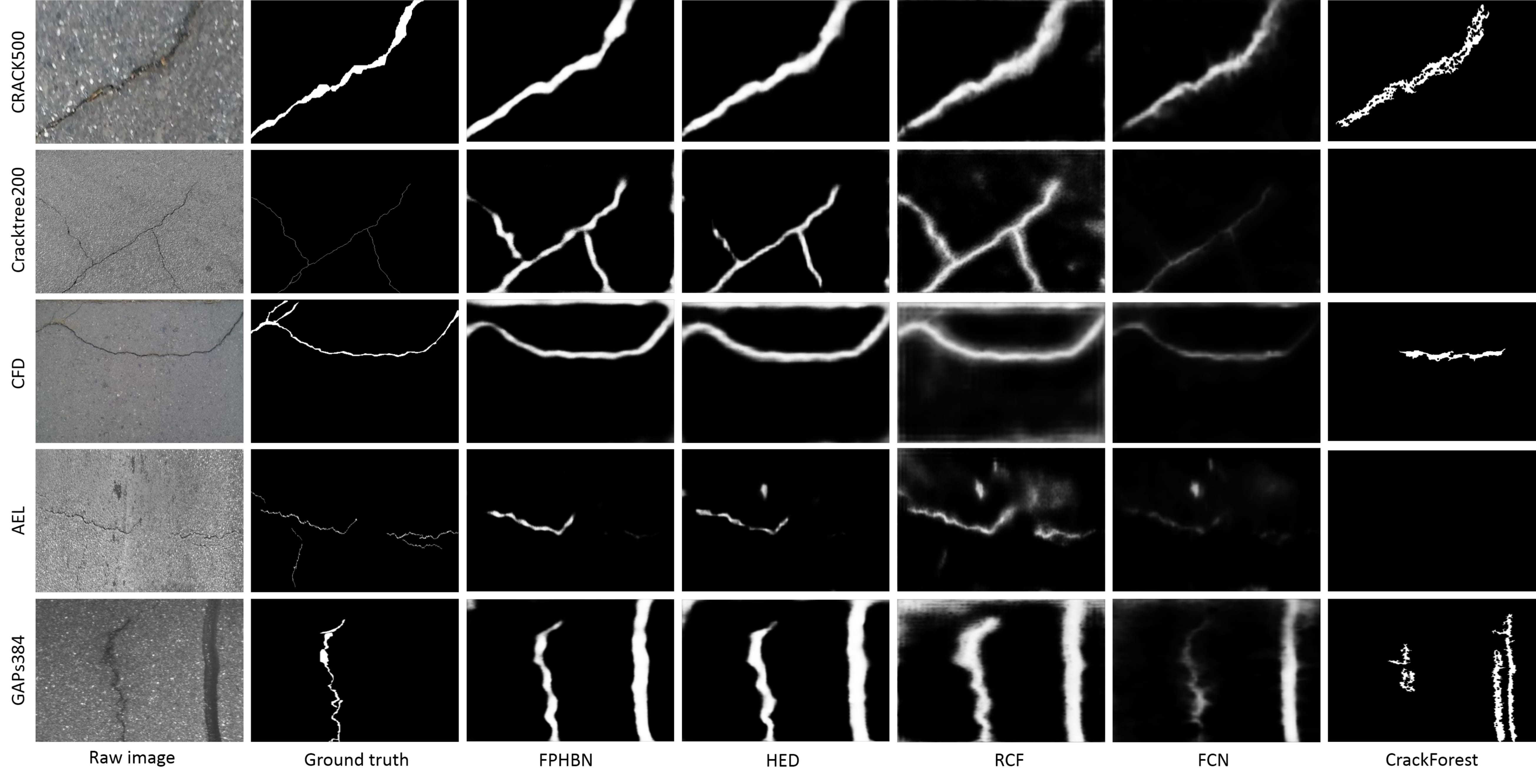}
	\caption{The visualization of detection results of compared methods on five datasets.}
	\label{fig:visualtest}
\end{figure*}
\begin{table}[t]
	\centering
	\small
	\caption{The AIU, ODS, and OIS of compared methods on CRACK500 test dataset.}	
	\begin{tabular}{c|c|c|c|c}
		\hline\hline
		Methods& AIU &ODS& OIS&time/image (s)\\
		\hline
		HED\cite{xie2015holistically} & 0.481 &0.575  & 0.625& 0.067 (GPU) \\ \hline
		RCF \cite{liu2017richer} & 0.403  &0.490  & 0.586 &\textbf{0.066 (GPU)}\\ \hline
		FCN\cite{long2015fully}  & 0.379 &0.513  & 0.577 & 0.101 (GPU)\\ \hline
		CrackForest \cite{shi2016automatic}  & N/A &0.199& 0.199 &2.951 (CPU)\\ \hline
		\textbf{FPHBN}& \textbf{0.489} & \textbf{0.604 } &  \textbf{0.635}  &0.197 (GPU) \\ \hline
	\end{tabular}
	
	\label{table:crack500test}
	
\end{table}

 \subsection{Experimental results}
  \subsubsection{Results on CRACK500}
  On CRACK500, we use the validation data to select the best training iteration. From Fig. \ref{fig:AIUval}, we can note that the AIU metric tends to converge over the training iterations. Based on the curve, we choose the model when its AIU curve has converged. In other word, FPHBN, HED\cite{xie2015holistically}, and RCF\cite{liu2017richer} are chosen at 12,000th iteration, FCN\cite{long2015fully} is chosen at 36,000th iteration. From TABLE \ref{table:crack500val}, we see that on validation dataset FPHBN surpasses RCF\cite{liu2017richer}, HED\cite{xie2015holistically}, and FCN\cite{long2015fully} in terms of AIU.

  We explore the contributions of each component of the proposed method on the validation set of CRACK500. As shown in TABLE I, we first introduce Feature Pyramid to HED, i.e., HED-FP. Compared with the original HED, we observe that the AIU improves from 0.541 to 0.553. We then integrate the Hierarchical Boosting into HED-FP, i.e., FPHBN. The AIU increases from 0.553 to 0.560 compared with HED-FP. Both the Feature Pyramid and Hierarchical Boosting contribute to the improvement of the performance.

  On the Crack500 test dataset, the evaluation curves and detection results of compared methods are shown in the first column of Fig. \ref{fig:curve} and the first row of Fig. \ref{fig:visualtest}, respectively.
  For intersection over union (IU) curve, we see that FPHBN always has considerably promising IU over various thresholds. For precision and recall (PR) curve, FPHBN is the highest among all compared methods. Since the output of CrackForest \cite{shi2016automatic} is a binary map, we cannot compute the IU and PR curve and just list the ODS and OIS value in related tables.

   As shown in TABLE \ref{table:crack500test}, FPHBN improves the performance by $5\%$ relative to HED\cite{xie2015holistically}, second best, in terms of ODS. In the first row of Fig. \ref{fig:visualtest}, we note that FPHBN gains visually much clearer crack detections than others. 
\begin{figure*}[t]
	\centering
	\includegraphics[width=1\textwidth,height=.35\linewidth]{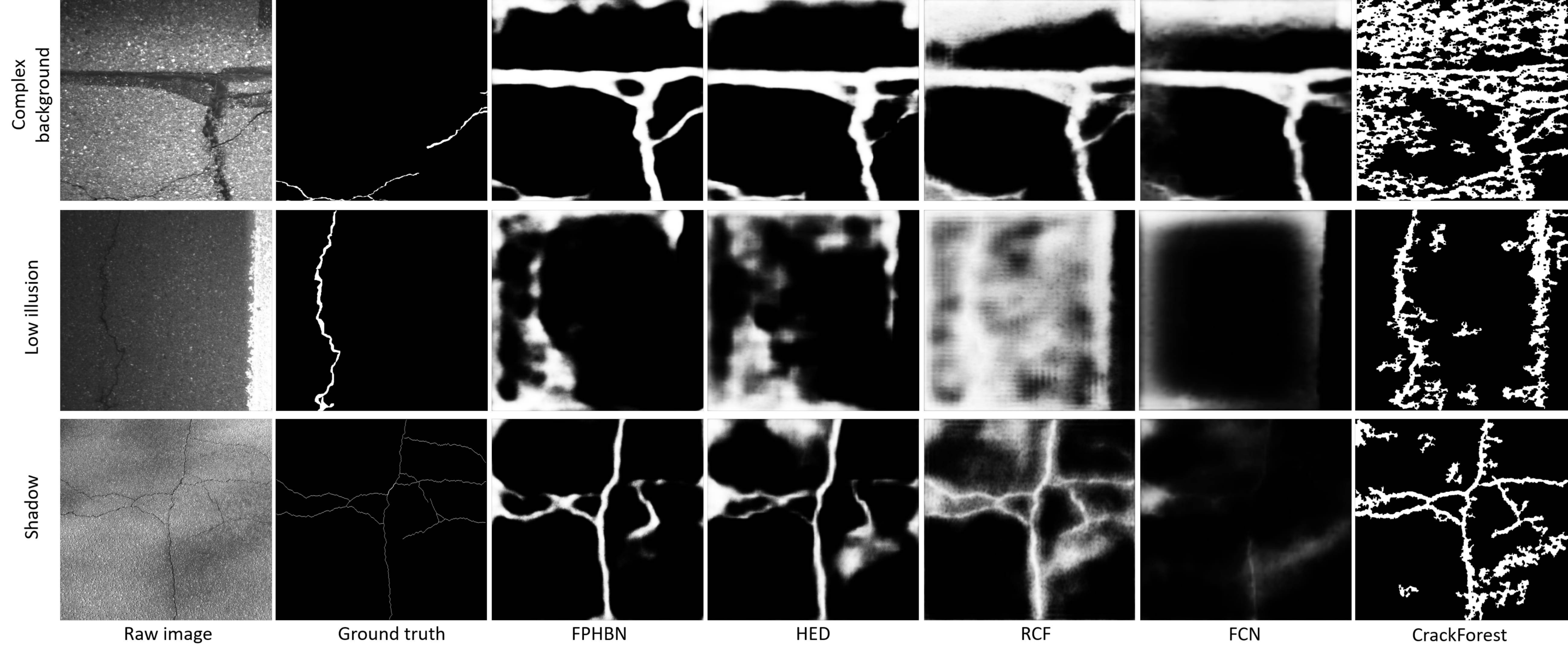}
	\caption{The visualization of detection results of compared methods on special cases, i.e., shadow, low illusion, and complex background.}
	\label{fig:visualspecial}
\end{figure*}
  \subsubsection{Results on GAPs384}

  \begin{table}[t]
  	\centering
  	\small
  	\caption{The AIU, ODS, and OIS of competing methods on GAPs384 dataset.}	
  	\begin{tabular}{c|c|c|c|c}
  		\hline\hline
  		Methods& AIU &ODS& OIS&time/image (s) \\
  		\hline
  		HED\cite{xie2015holistically} &0.069 & 0.209 & 0.175& 0.086 (GPU)\\ \hline
  		RCF \cite{liu2017richer} & 0.043 &0.172  & 0.120  & \textbf{0.083 (GPU)}\\ \hline
  		FCN\cite{long2015fully}  & 0.015 &0.088  & 0.091  &0.119 (GPU)\\ \hline
  		CrackForest \cite{shi2016automatic}  &N/A &0.126 & 0.126  &4.101 (CPU)\\ \hline
  		\textbf{FPHBN}  & \textbf{0.081} & \textbf{0.220} &  \textbf{0.231} & 0.241 (GPU)\\ \hline
  	\end{tabular}
  	\label{table:GAPs}
  \end{table}
  \begin{table}[t]
  	\centering
  	\small
  	\caption{The AIU, ODS, and OIS of compared methods on Cracktree200 dataset.}	
  	\begin{tabular}{c|c|c|c|c}
  		\hline\hline
  		Methods& AIU&ODS& OIS &time/image (s)\\
  		\hline
  		HED\cite{xie2015holistically} & 0.040 & 0.317 & 0.449 &0.130(GPU)\\ \hline
  		RCF\cite{liu2017richer}  & 0.032 &0.255   & 0.487   &\textbf{0.128 (GPU)}\\ \hline
  		FCN\cite{long2015fully}  & 0.008 &0.334  & 0.333   &0.166 (GPU)\\ \hline
  		CrackForest  \cite{shi2016automatic} & N/A &0.080 &0.080  &5.091 (CPU)\\ \hline
  		\textbf{FPHBN}  & \textbf{0.041}& \textbf{0.517} &  \textbf{0.579} &0.377 (GPU)\\ \hline
  	\end{tabular}
  	
  	\label{table:cracktree200}
  	
  \end{table}
  \begin{table}[t]
  	\centering
  	\small
  	\caption{The AIU, ODS, and OIS of competing methods on CFD dataset.}	
  	\begin{tabular}{c|c|c|c|c}
  		\hline\hline
  		Methods& AIU &ODS& OIS &time/image (s)\\
  		\hline
  		HED\cite{xie2015holistically} & 0.154&0.593  & 0.626 &0.047 (GPU)\\ \hline
  		RCF\cite{liu2017richer}  & 0.105 &0.542  & 0.607    &\textbf{0.040 (GPU)}\\ \hline
  		FCN\cite{long2015fully}  & 0.021 &0.585  & 0.609  &0.07 (GPU)\\ \hline
  		CrackForest \cite{shi2016automatic}  & N/A &0.104& 0.104  &3.742 (CPU)\\ \hline
  		\textbf{FPHBN}  & \textbf{0.173} & \textbf{0.683} & \textbf{0.705}    &0.133 (GPU)\\ \hline
  	\end{tabular}
  	\label{table:cfd}
  \end{table}
  \begin{table}[t]
  	\centering
  	\small
  	\caption{The AIU, ODS, and OIS of competing methods on AEL dataset.}	
  	\begin{tabular}{c|c|c|c|c}
  		\hline\hline
  		Methods& AIU &ODS& OIS&time/image (s) \\
  		\hline
  		HED\cite{xie2015holistically} & 0.075&0.429  & 0.421 &0.098 (GPU)\\ \hline
  		RCF\cite{liu2017richer}  & 0.069 &0.469  & 0.397    &\textbf{0.097 (GPU)}\\ \hline
  		FCN\cite{long2015fully}  & 0.022 &0.322  & 0.265  &0.128 (GPU)\\ \hline
  		CrackForest \cite{shi2016automatic}  & N/A &0.231  & 0.231  &2.581 (CPU)\\ \hline
  		\textbf{FPHBN}  & \textbf{0.079} & \textbf{0.492} &  \textbf{0.507}    &0.259 (GPU)\\ \hline
  	\end{tabular}
  	\label{table:ael}
  \end{table}
  \begin{table}[t]
  	\centering
  	\small
  	\caption{The mean and standard deviation (std) of ODS and OIS over datasets GAPs384, Cracktree200, CFD, and AEL.}
  	\label{table:generality}
  	\begin{tabular}{c|c|c}
  		\hline
  		\hline
  		Method &  ODS (mean$\pm$std)&OIS (mean$\pm$std)\\ \hline
  		HED\cite{xie2015holistically}& 0.387 $\pm$ 0.164&0.418 $\pm$  0.186 \\ \hline
  		RCF\cite{liu2017richer}&  0.359$\pm$0.175& 0.403 $\pm$  0.207 \\ \hline
  		FCN\cite{long2015fully}&  0.332$\pm$0.203&  0.325 $\pm$  0.215\\ \hline
  		CrackForest \cite{shi2016automatic}&  0.135$\pm$0.067&0.135$\pm$0.067\\ \hline
  		\textbf{FPHBN}&  \textbf{0.478}$\pm$0.192&  \textbf{0.505} $\pm$  0.201\\ \hline
  	\end{tabular}
  \end{table}
  From Fig. \ref{fig:curve} and TABLE \ref{table:GAPs}, we see that the proposed FPHBN achieves best performance. However, the gained performance is much lower than that on other datasets. It is because the GAPs384 dataset has non-uniform illumination and similar background. For example, in the last row of Fig. \ref{fig:visualtest}, a sealed crack is by the side of a true crack and misclassified as a crack.
  In addition, we note that the AIU value of all compared methods are very small. This is because the ground truth of GAPs384 is one or several pixels wide.
  \subsubsection{Results on Cracktree200}

  From the second column of Fig. \ref{fig:curve} and the second row of Fig. \ref{fig:visualtest}, we can see that FPHBN  gains best performance. Especially, in PR curve, FPHBN improves the performance by a large margin. In TABLE \ref{table:cracktree200}, we see that FPHBN outperforms HED by $63.1\%$ and $28.9\%$  in ODS and OIS, respectively.
  \subsubsection{Results on CFD}
  From the third column of Fig. \ref{fig:curve}, we see that FPHBN achieves superior performance compared to HED\cite{xie2015holistically}, RCF\cite{liu2017richer}, and FCN\cite{long2015fully}. In TABLE \ref{table:cfd}, FPHBN improves HED \cite{xie2015holistically} by $15.2\%$ and $12.6\%$ in terms of ODS and OIS, respectively. 
  \subsubsection{Results on AEL}
  As shown in Fig. \ref{fig:curve} and TABLE \ref{table:ael}, although RCF\cite{liu2017richer} gains the highest value on IU curve, the best AIU is achieved by FPHBN. From Fig. \ref{fig:visualtest}, we note that FPHBN has much less false positives than the other methods. In TABLE \ref{table:ael}, FPHBN increases ODS and OIS by $4.9\%$ and $20.4\%$ compared with the second best, respectively.

 \subsection{Cross dataset generalization}
  To compare the generalizability of compared methods, we compute mean and standard deviation of the ODS and OIS over datasets GAPs384, Cracktree200, CFD, and AEL. TABLE.\ref{table:generality} shows the quantitative results. We note that the proposed FPHBN achieves best mean ODS and OIS and surpasses the second best by a large margin. This demonstrates that the proposed method has significantly better generalizability than state-of-the-art methods.  The reason of the superior performance of the proposed method can be attributed in two aspects: 1. multi-scale context information is fed into low-level layers via the top-down feature pyramid structure to enrich the feature representation in low-level layers for crack detection; and 2. hard example mining is performed by the hierarchical boosting which helps the side networks to perform crack detection in a complementary way. The low-level side networks focus on pixels that are not well classified by high-level ones,  such that the final detection performance is improved.

  \subsection{Speed comparison}
   We test the inference time of compared methods on all datasets. Since  CrackForest\cite{shi2016automatic} is not implemented on GPU, only CPU time is listed. From TABLE \ref{table:crack500val} to \ref{table:ael}, we note that the proposed FPHBN is slower than HED \cite{xie2015holistically} around $0.086$s to $0.249$s. This is because the feature pyramid increases the computation expense for each side network. Although our method is not real time, the speed can be improved by the development of hardware and the technology of model compression \cite{han2015deep}.

 \subsection{Special cases discussion}
 To further compare and analyze the proposed and state-of-the-art methods, we conduct experiments on some special cases, i.e., complex background, low illumination, and shadow. Fig. \ref{fig:visualspecial} shows representative results of all methods. In complex background, the real crack is surrounded by sealed crack. In this case, all algorithms misclassify the sealed crack as crack. Compared with the state-of-the-art methods, the proposed FPHBN yields a clearer and better result (the first row of Fig. \ref{fig:visualspecial}). The reason of the failure of these methods is the similar pattern between crack and sealed crack.

 In low illumination condition, all methods fail to detect crack. This is because these scenarios are unseen in training data. An appropriate data augmentation may solve the problem to certain extent. For the scene with shadow, compared with HED \cite{xie2015holistically}, the proposed FPHBN produces much fewer false positives. This indicates that FPHBN is more robust to shadows than HED \cite{xie2015holistically}, which can be contributed to the feature pyramid and hierarchical boosting.
\section{Conclusion}
In this work, a feature pyramid and hierarchical boosting network (FPHBN) is proposed for pavement crack detection. The feature pyramid is introduced to enrich the low-level feature by integrating semantic information from high-level layers in a pyramid way. A hierarchical boosting module is proposed to deal with hard examples by reweighting samples in a hierarchical way. Incorporating the two components to HED\cite{xie2015holistically} results in the proposed FPHBN. A novel crack detection measurement, i.e., AIU has been proposed.  Extensive experiments are conducted to demonstrate the superiority and generalizability of FPHBN.
\label{conclusion}


%






\bibliographystyle{IEEEtran}
\bibliography{tits_crack}
%


%


\vspace{-0.7in}
\begin{IEEEbiography}[{\includegraphics[width=25.4 mm,height=25.4 mm]{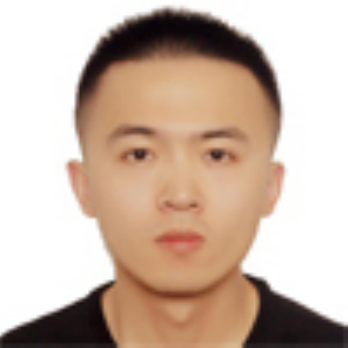}}]{Fan Yang}
	received the B.S. degree in Electrical Engineering from Anhui Science and Technology University, Bengbu, and M.S. degree in Biomedical Engineering from Xi'dian University, Xi'an, China, in 2012 and 2015, respectively. He is currently a  Ph.D student in Computer and Information Sciences Department at Temple University. His current research interests are computer vision and machine learning.
\end{IEEEbiography}
\vspace{-0.7in}
\begin{IEEEbiography}[{\includegraphics[width=25 mm,height=25.4 mm]{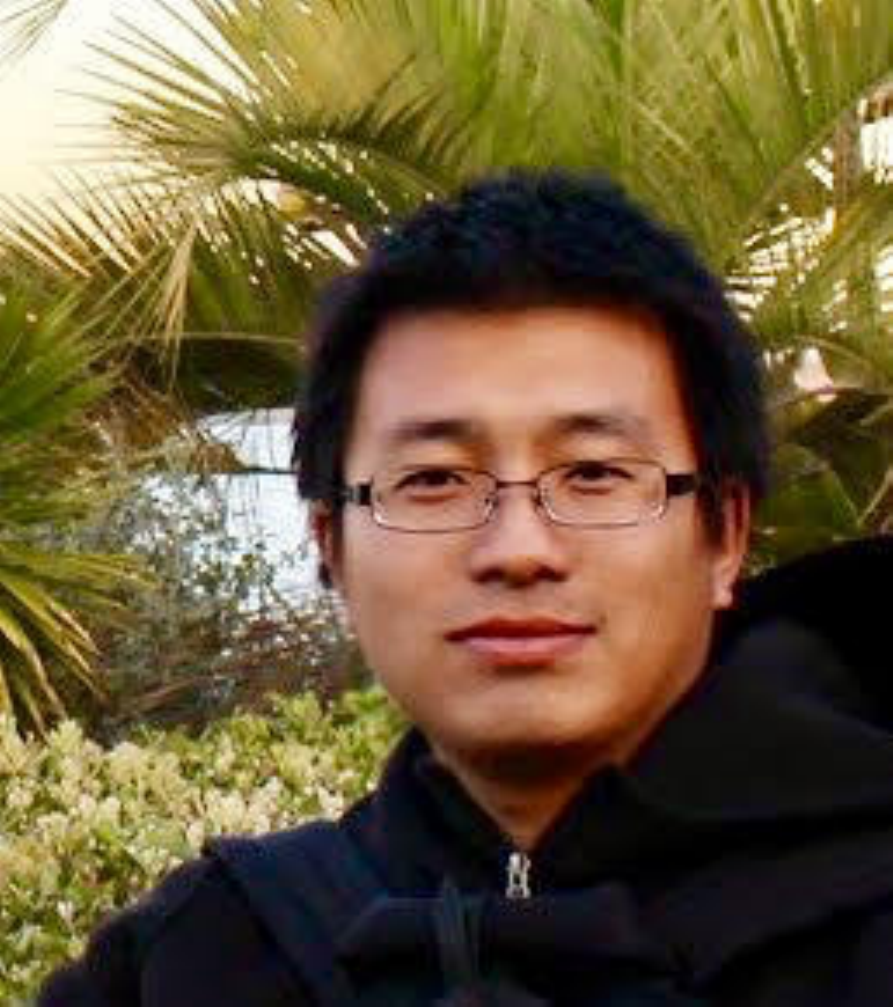}}]{Lei Zhang}
	 received the Ph.D. degree in computer
	 science from Harbin Institute of Technology, Harbin,
	 China, in 2013. From September 2011 to August
	 2012, He was a research intern in Siemens Corporate
	 Research, Inc., Princeton, NJ. From July
	 2015 to September 2017, He was a Post-Doctoral
	 Research Fellow in College of Engineering, Temple
	 University, PA and a Post-Doctoral Research Fellow
	 in Department of Biomedical Informatics, Arizona
	 State University, Scottsdale, AZ, respectively. He is
	 currently a Postdoctoral associate in the School of
	 Medicine, Department of Radiology at the University of Pittsburgh. He was
	 a lecturer in School of Art and Design, Harbin University, Harbin, China.
	 His current research interests include machine learning, computer vision,
	 visualization and medical image analysis.
	
\end{IEEEbiography}
\vspace{-0.7in}
\begin{IEEEbiography}[{\includegraphics[width=25.4 mm,height=25.4 mm]{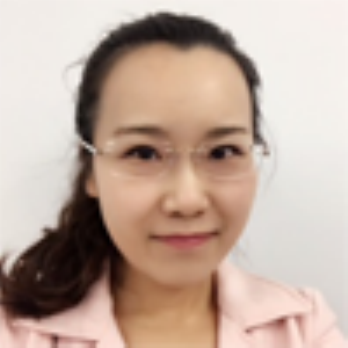}}]{Sijia Yu}	
	received the B.S. degree in Biotechnology from China Pharmaceutical University, Nanjing, China, and M.S. degree in Biomedical Science from Temple University, Philadelphia, USA. She is currently a  Master's student in Computer add Information Sciences Department at Temple University and she is working in Dr. Ling Haibin's Lab at Temple University. Her current research interests is computer vision.
	
\end{IEEEbiography}
\vspace{-0.7in}
\begin{IEEEbiography}[{\includegraphics[width=25.4 mm,height=25.4 mm]{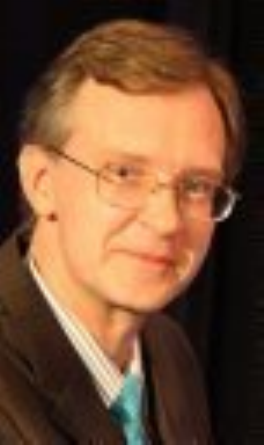}}]{Danil Prokhorov}
was a Research Engineer with the St. Petersbu3rg Institute for Informatics and Automation, Russian Academy of Sciences, Saint Petersburg, Russia. He has been involved in automotive research since 1995. He was an Intern with the Scientific Research Laboratory, Ford Motor Company, Dearborn, MI, USA, in 1995. In 1997, he became a Research Staff Member with Ford Motor Company, where he was involved in application-driven research on neural networks and other methods. Since 2005, he has been with Toyota Technical Center, Ann Arbor, MI, USA. He is currently in charge of Department of Future Mobility Research, Toyota Research Institute, Ann Arbor.

\end{IEEEbiography}
\vspace{-0.7in}
\begin{IEEEbiography}[{\includegraphics[width=25.4 mm,height=25.4 mm]{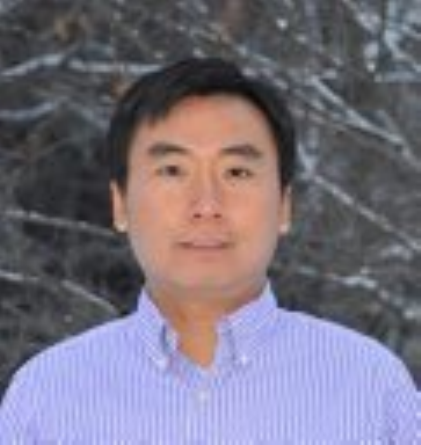}}]{Xue Mei}
received the B.S. degree from the University of Science and Technology of China, Hefei, China, and the Ph.D. degree from the University of Maryland, College Park, College Park, MD, USA, both in electrical engineering. He is a Senior Research Scientist with the Department of Future Mobility Research, Toyota Research Institute, Ann Arbor, MI, USA, a Toyota Technical Center Division. From 2008 to 2012, he was with Automation Path-Finding Group in Assembly and Test Technology Development and Visual Computing Group with Intel Corporation, Santa Clara, CA, USA. His current research interests include computer vision, machine learning, and robotics with a focus on intelligent vehicles research.
\end{IEEEbiography}
\vspace{-0.6in}
\begin{IEEEbiography}[{\includegraphics[width=25.4 mm,height=25.mm]{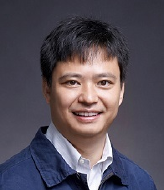}}]{Haibin Ling}
received the B.S. degree in mathematics and the MS degree in computer science from Peking University, China, in 1997 and 2000, respectively, and the PhD degree from the University of Maryland, College Park, in Computer Science in 2006. From 2000 to 2001, he was an assistant researcher at Microsoft Research Asia. From 2006 to 2007, he worked as a postdoctoral scientist at the University of California Los Angeles. After that, he joined Siemens Corporate Research as a research scientist. Since fall 2008, he has been with Temple University where he is now an Associate Professor. His research interests include computer vision, medical image analysis and machine learning.

\end{IEEEbiography}
\end{document}